\documentclass{article}

\usepackage{arxiv}

\usepackage[utf8]{inputenc} % allow utf-8 input
\usepackage[T1]{fontenc}    % use 8-bit T1 fonts
\usepackage{hyperref}       % hyperlinks
\usepackage{url}            % simple URL typesetting
\usepackage{booktabs}       % professional-quality tables
\usepackage{amsfonts}       % blackboard math symbols
\usepackage{nicefrac}       % compact symbols for 1/2, etc.
\usepackage{microtype}      % microtypography
\usepackage{lipsum}		% Can be removed after putting your text content
\usepackage{graphicx}
\usepackage{natbib}
\usepackage{doi}
\usepackage{multirow}
\usepackage{tablefootnote}

\title{Text2KGBench: A Benchmark for Ontology-Driven Knowledge Graph Generation from Text}

\author{ \href{https://orcid.org/0000-0003-1707-4842}{\includegraphics[scale=0.06]{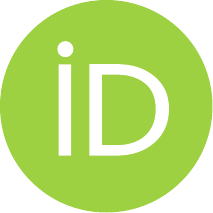}\hspace{1mm}Nandana Mihindukulasooriya}\thanks{Corresponding author.} \\
	IBM Research Europe \\
	Ireland\\
	\texttt{nandana@ibm.com} \\
	\And
	\href{https://orcid.org/0000-0001-7197-0766}{\includegraphics[scale=0.06]{orcid.pdf}\hspace{1mm}Sanju Tiwari} \\
	Universidad Autonoma de Tamaulipas, Mexico\\
		\texttt{tiwarisanju18@ieee.org} \\
        \And
	\href{https://orcid.org/0000-0003-4303-983X}{\includegraphics[scale=0.06]{orcid.pdf}\hspace{1mm}Carlos F. Enguix} \\
    Universidad Autonoma de Tamaulipas, Mexico \\
		\texttt{carlos.f.enguix@gmail.com} \\
        \And
	\href{https://orcid.org/0000-0002-9492-7653}{\includegraphics[scale=0.06]{orcid.pdf}\hspace{1mm}Kusum Lata} \\
    Sharda University, India\\
		\texttt{kusumlata.1@sharda.ac.in} \\
}

\renewcommand{\shorttitle}{Text2KGBench: A Benchmark for Ontology-Driven KG Generation from Text}

\hypersetup{
pdftitle={Text2KGBench},
pdfsubject={q-bio.NC, q-bio.QM},
pdfauthor={Nandana Mihindukulasooriya1, Sanju
Tiwari, Carlos F. Enguix, and Kusum Lata},
pdfkeywords={Benchmark, Relation Extraction, Knowledge Graph, Knowledge Graph Generation, Large Language Models }
}

\begin{document}
\maketitle

\begin{abstract}
	The recent advances in large language models (LLM) and foundation models with emergent capabilities have been shown to improve the performance of many NLP tasks. LLMs and Knowledge Graphs (KG) can complement each other such that LLMs can be used for KG construction or completion while existing KGs can be used for different tasks such as making LLM outputs explainable or fact-checking in Neuro-Symbolic manner. In this paper, we present \textit{Text2KGBench}, a benchmark to evaluate the capabilities of language models to generate KGs from natural language text guided by an ontology. Given an input ontology and a set of sentences, the task is to extract facts from the text while complying with the given ontology (concepts, relations, domain/range constraints) and being faithful to the input sentences. We provide two datasets (i) Wikidata-TekGen with 10 ontologies and 13,474 sentences and (ii) DBpedia-WebNLG with 19 ontologies and 4,860 sentences. We define seven evaluation metrics to measure fact extraction performance, ontology conformance, and hallucinations by LLMs. Furthermore, we provide results for two baseline models, Vicuna-13B and Alpaca-LoRA-13B using automatic prompt generation from test cases. The baseline results show that there is room for improvement using both Semantic Web and Natural Language Processing techniques.
 \\\\
\textbf{Resource Type:} Evaluation Benchmark\\
~\textbf{Source Repo:} \url{https://github.com/cenguix/Text2KGBench} \\
~\textbf{DOI:} \url{https://doi.org/10.5281/zenodo.7916716} \\
~\textbf{License:  Creative Commons Attribution (CC BY 4.0)} 
\end{abstract}

% keywords can be removed
\keywords{Benchmark  \and Relation Extraction \and Knowledge Graph \and Knowledge Graph Generation \and Large Language Models}

\section{Introduction}
Knowledge Graphs (KG) are becoming popular in both industry and academia due to their useful applications in a wide range of tasks such as question answering, recommendations, semantic search, and advanced analytics with explainability \citep{hogan2021knowledge}. A KG can be generated using mappings such as RDB2RDF \citep{sahoo2009survey} if the source is relational data or semi-structured using RML \citep{dimou2014rml}. Crowdsourcing can be used to build them manually as in Wikidata \citep{vrandevcic2014wikidata}. However, there are cases where the data is in unstructured format in text documents and crowd-sourcing is not an option (for example, internal documents). One solution in such cases is to construct knowledge graphs using Natural Language Processing (NLP) techniques such as Named Entity Recognition (NER), Relation Extraction, Open Information Extraction, Entity Linking, and Relation Linking. There is a growing interest in the Semantic Web community to explore such approaches as seen from the workshops such as Text2KG \citep{2022text2kg,2023text2kg} and NLP4KGC \citep{vakaj2023nlp4kgc}.

The recent advances in large language models (LLM) and foundation models with emergent capabilities have been shown to improve the performance in many NLP tasks \citep{brown2020language}. KGs and LLMs can complement each other in both directions; on the one hand, LLMs can be helpful in constructing KGs and on the other hand KGs can be used to validate LLM outputs or make them explainable. Approaches such as Neuro-Symbolic AI \citep{hitzler2022neuro} will allow using KGs and LLMs jointly. In order to foment research in this direction, the establishment of evaluation benchmarks is necessary. In this context, \textit{Text2KGBench} is a benchmark for measuring the capabilities of LLMs for generating KGs from text conforming to a given ontology. In this version, we are not evaluating the ability to process or generate RDF/OWL representations but rather the ability of extracting facts using correct relations.

There are several manners LLMs can be adapted to this task, including fine tuning \citep{fine-tuning-paper} (also known as model tuning), updating all model parameters, Prompt tuning \citep{lester-etal-2021-power} or Prefix-Tuning \citep{Li2021PrefixTuningOC} by keeping the model parameters frozen and only prefixing some tunable tokens to the input text and prompt design where the model is used as it is, but the prompt or the input to the model is designed to provide a few examples of the task \citep{brown2020language}. Each of these approaches has their pros and cons with respect to the performance, computation resources, training time, domain adaption and training data required. Our benchmark provides training data that can be used in any of those approaches.

In-context learning \citep{min-etal-2022-rethinking,xie2021explanation} with prompt design is about teaching a model to perform a new task only by providing a few demonstrations of input-output pairs at inference time. Instruction fine-tuning using approaches such as InstructGPT \citep{ouyang2022training}, Reinforcement Learning from Human Feedback (RLHF) \citep{christiano2017deep,stiennon2020learning} significantly improves the models capabilities to follow a broad range of written instructions.

A vast number of LLMs have been released in recent months \citep{yang2023harnessing}, especially in the GPT family of models such as GPT-3 \citep{brown2020language}, ChatGPT, LLaMA \citep{touvron2023llama}, BLOOM \citep{scao2022bloom}, PaLM \citep{chowdhery2022palm}, and Bard. Such models can be easily adapted for KG generation from text with a prompt design containing instructions and contextual information.

The main contributions of this paper are:
\begin{itemize}
    \item We propose a novel benchmark \textit{Text2KGBench} by extending the relation extraction by guiding it with ontology and instructions. We provide two datasets, (a) Wikidata-TekGen with 10 ontologies and 13,474 sentences aligned to triples and (b) DBpedia-WebNLG with 19 ontologies and 4,860 sentences aligned to triples by reusing TekGen \citep{agarwal-etal-2021-knowledge} and WebNLG \citep{gardent2017creating} corpora.
    We define seven metrics for measuring the accuracy of fact extraction, ontology conformance and detecting hallucinations and provide evaluation scripts.   
    \item We provide results for two baselines using open-source LLMs, including Vicuna-13B \citep{vicuna2023} and Alpaca-LoRA-13B \citep{alpaca,hu2022lora} with in-context learning. We also provide a baseline automatic prompt generator from ontologies and approach finding best demonstration examples with sentence similarity using SBERT T5-XXL model \citep{reimers-2019-sentence-bert,ni-etal-2022-large}. We provide all generated prompts, similarities, and LLM responses for further analysis.
\end{itemize}

The rest of the paper is organized as follows. Section~\ref{sec.task_desc} introduces the task of the benchmark, Section~\ref{sec.bench.gen} describes how the benchmark was created, Section~\ref{sec.eval.metrics} defines the evaluation metrics and Section~\ref{sec.eval.results} presents the baselines and evaluation results. After related work in Section~\ref{sec.related.work}, the paper concludes with some final remarks and future work in Section~\ref{sec.conclusions}.

\section {Task Description}
\label{sec.task_desc}
\begin{figure}[h!]
\centering
\includegraphics[width=1\textwidth]{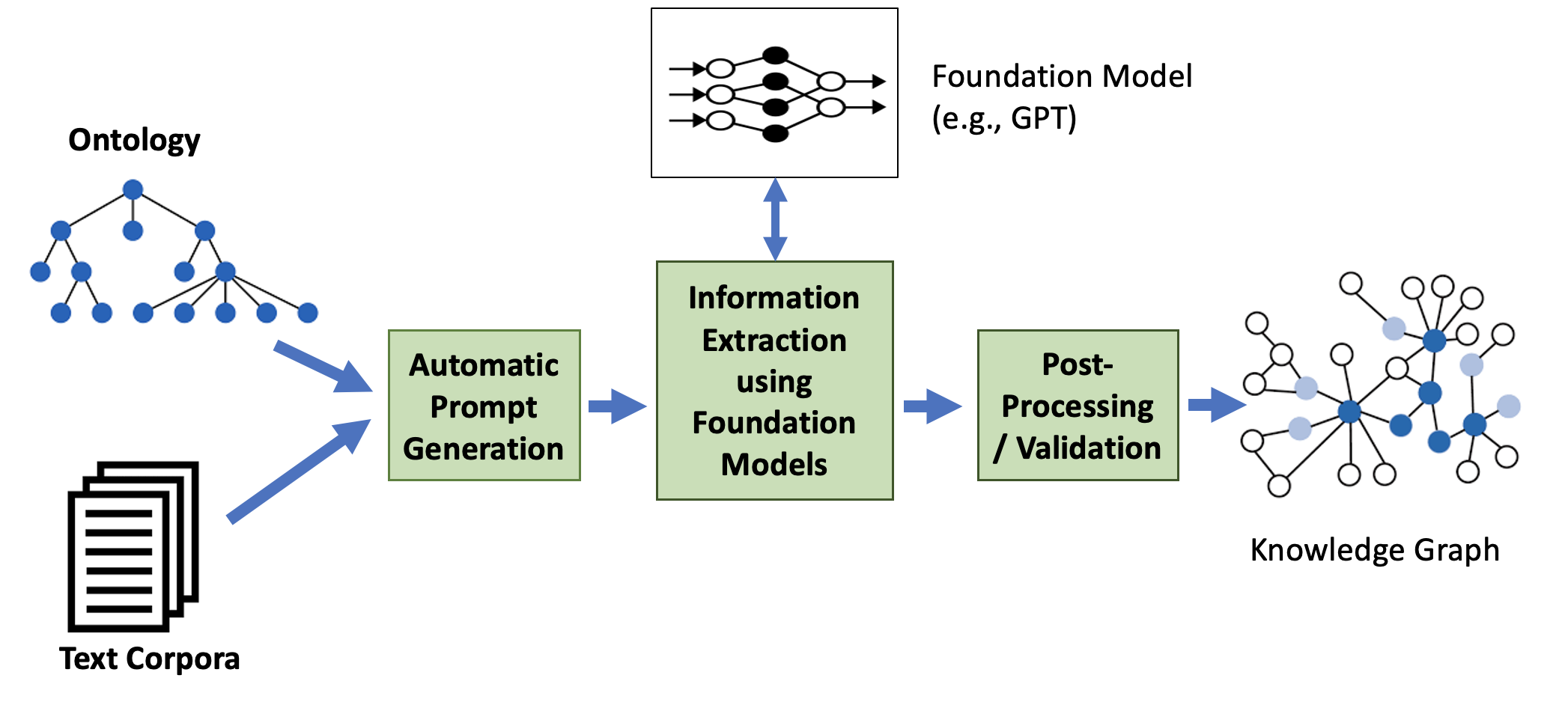}
\caption{Generating knowledge graphs from text guided by ontologies}
\label{fig:task}
\end{figure}

This section introduces the task of \textit{Text2KGBench}. With the recent advancements of LLMs, we envision that LLMs can be used to generate KGs guided by ontologies as illustrated in Figure~\ref{fig:task}. Given an ontology and text corpora, the goal is to construct prompts to instruct the model to extract facts relevant to the ontology. Such extracted facts can be further validated and post-processed to create a knowledge graph. 

In the context of the \textit{Text2KGBench}, we define the task as a fact extraction task guided by an ontology. The proposed task is closely related to the relation extraction and relation classification tasks in literature but with an explicit ontology definition given as input. There are three main inputs to the task:\\
\textit{Ontology}: The ontology defines the concepts of interest, a set of defined relations with their canonical names, domain and range constraints for the relations. This can be further extended with other ontological axioms to guide models. \\
\textit{Text Corpus}: The text corpus contains the set of natural language sentences that contains facts that can be expressed using the aforementioned ontology. \\
\textit{Examples}: Demonstrative examples or training data contains pairs of sentences and the facts extracted from them complying with the ontology. 

Given these inputs, a system should be able to generate facts adhering to a set of expectations. First, the system should use the ontology and demonstrative examples as guidance on which facts to extract and which relations to be used in the output. It should follow the canonical relation names and the example output format. In the evaluation, we measure this aspect using ontology compliance metrics. Second, the system should be faithful to the input sentence. This means the system should consider only the facts mentioned in the sentence as the truth (irrespective of the knowledge it may have from pre-training). It should not include additional information that is not directly or indirectly stated or implied by the sentence. This aspect is measured by the fact extraction accuracy metrics. Finally, the system should not hallucinate i.e. it should not introduce new or fake entities/relations not mentioned in the sentence and the ontology. This aspect is measured by the hallucination metrics. Section~\ref{sec.eval.metrics} provides details of evaluation metrics. 

\begin{figure}[h]
\centering
\includegraphics[width=1\textwidth]{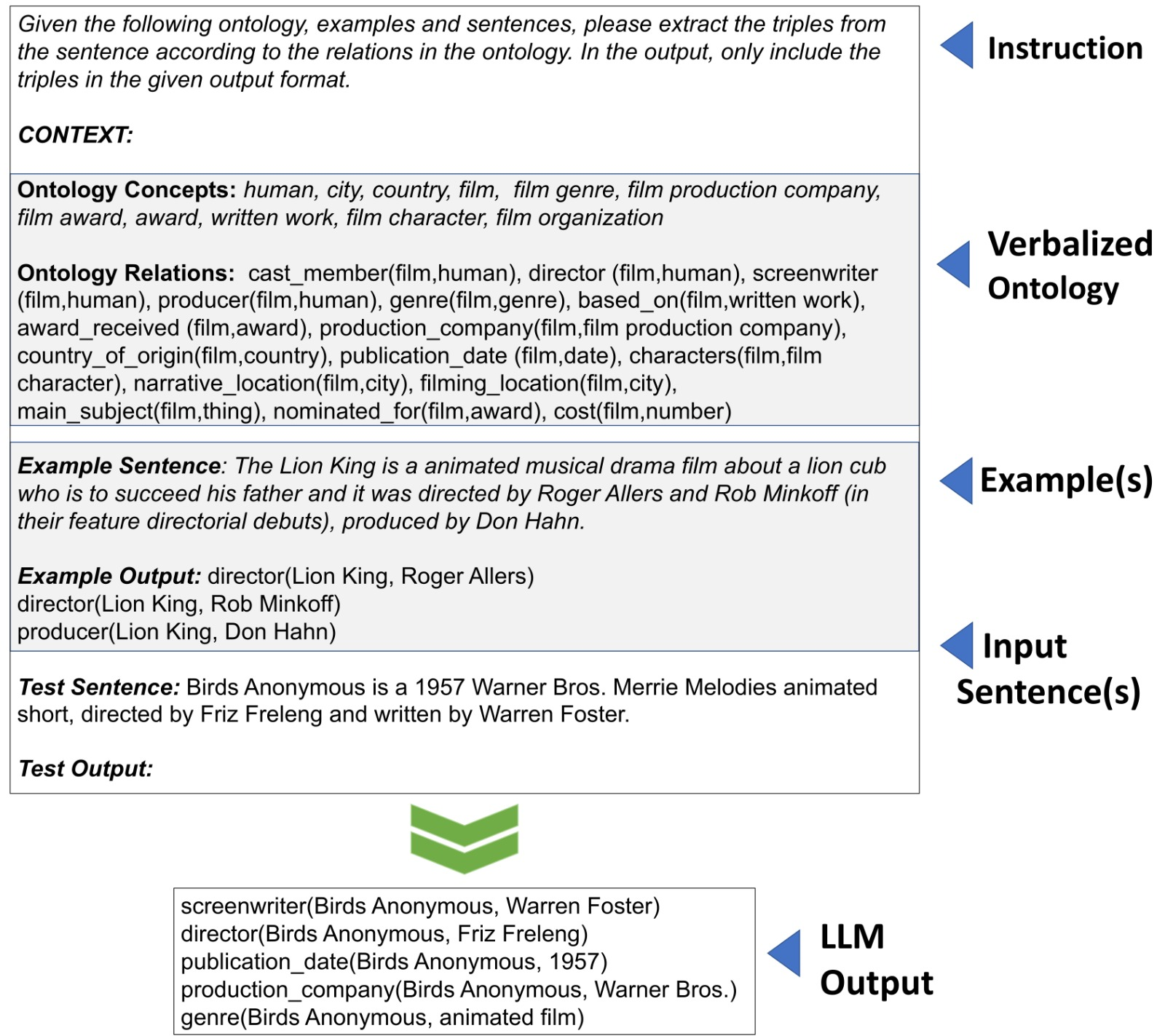
}
\caption{An example prompt for an instruction fine-tuned LLM and the generated output from the LLM model.}
\label{fig:prompt}
\end{figure}

In this version of \textit{Text2KGBench}, we are not evaluating a system's ability to process RDF/OWL syntax or a deep understanding of OWL semantics. Thus, we are using simpler language-oriented verbalizations and triple formats for presenting the information to an LLM. Figure~\ref{fig:prompt} illustrates an example of performing the task using in-context learning of LLMs with a prompt.

There are several components or lines of research that can affect the results of a system tested under this benchmark. One of the most important aspects is the model (LLM) being used. Depending on the characteristics such as the architecture, training data being used, number of parameters, and what instructions have been used for fine-tuning, each of the language models can have different capabilities, and it has a direct impact on the results obtained from the model. 

Prompt engineering or automatic prompt generation also plays a vital role in this task. Recently, there is a line of research that is focused on how to build efficient prompts for getting expected outputs from LLMs. In this benchmark, the participants can design different prompts guided by an ontology and reasoning techniques can be used to develop the most efficient prompts. Related to prompt generation, another important aspect is how to find the most relevant or helpful demonstration example from training data given a test case. This can be done using sentence similarity metrics or utilizing more advanced semantic clues from the ontology. 

Post-processing and validation are also crucial for extracting the correct triples and cleaning them by removing implausible triples. Initial extraction can be done using pattern-matching techniques such as using regex. Validation of the generated triples is another open research area which can use linguistic approaches to detect hallucinations and reasoning-based approaches to validate that the generated triples are consistent with the ontology.

\section {Benchmark Generation}
\label{sec.bench.gen}

\textit{Text2KGBench} consists of two datasets: \textit{wikidata-tekgen} and \textit{dbpedia-webnlg}. As discussed above, each of those has a set of ontologies and corpora of text where sentences are aligned with triples according to the given ontology.

\subsection{Wikidata-TekGen Dataset}

This dataset is created using sentence alignments provided by TekGen corpus.

\paragraph{\textbf{Ontology Selection}}
As the first step of building the dataset, we have created 10 small ontologies by reusing the concepts and relations described in Wikidata. We selected a domain, such as movies or sports and explored the concepts and relations relevant to the given domain in Wikidata. With that, a set of concepts for the domain are identified, and a sample of their instances is checked for the most frequent relations. Once a relation is identified, it's property page is used to understand the usage, and domain range constraints. For example, the property page\footnote{\url{https://www.wikidata.org/wiki/Property:P57}} for the relation ``director (P57)" describes subject and value-type constraints. Iteratively, more concepts are added to the ontology based on the domain/range constraints of the selected relations. This process is performed manually and each ontology was formulated by an author with Semantic Web expertise and reviewed by two other experts. Table~\ref{tab:onto_stats} shows the concept and relation statistics for each of the 10 ontologies that we generated.

\begin{figure}[h!]
\centering
\includegraphics[width=1\textwidth]{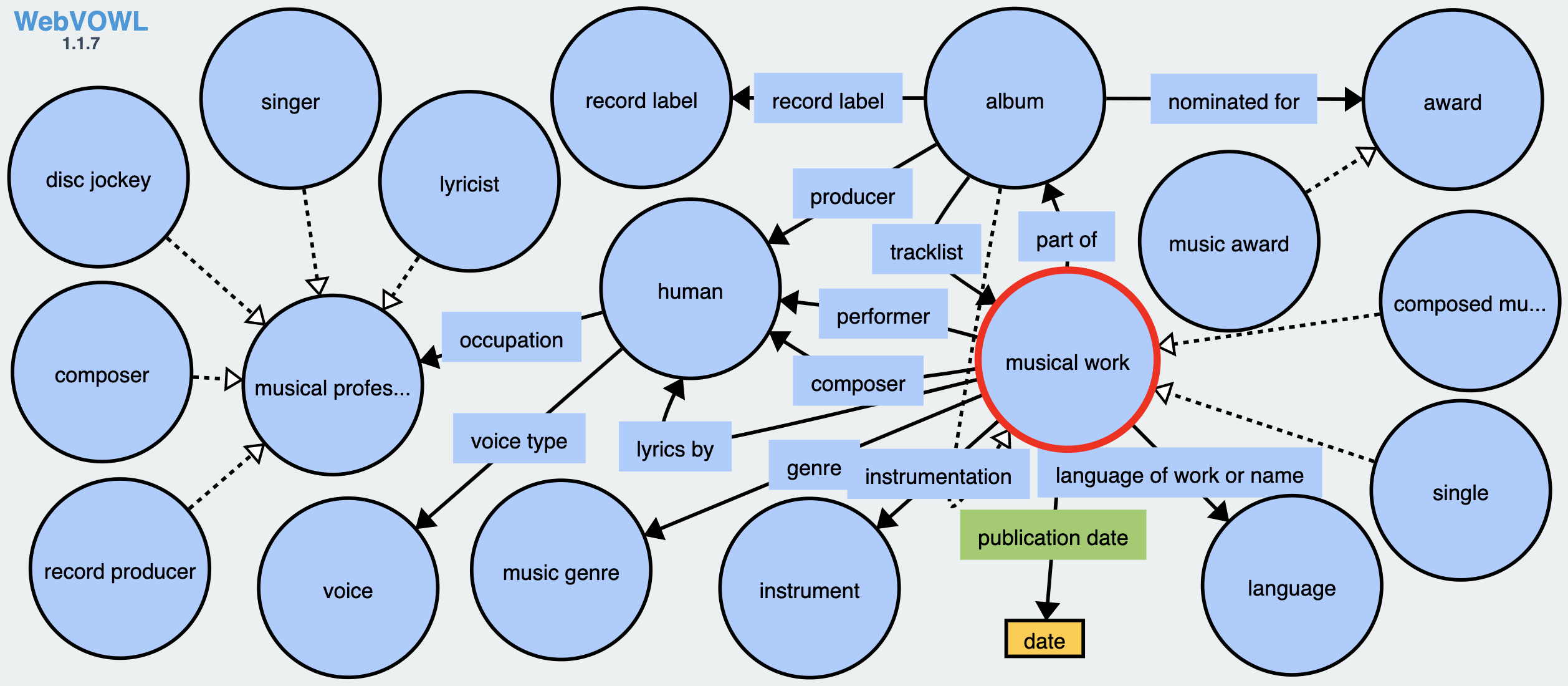}
\caption{An illustration of the music ontology with concepts and relations selected from Wikidata.}
\label{fig:music_ontology}
\end{figure}

An example ontology for the music domain is shown in Figure~\ref{fig:music_ontology}. All 10 ontologies are available as OWL ontologies serialized in Turtle and in a compact json format in the repo\footnote{\url{https://github.com/cenguix/Text2KGBench/tree/main/data/wikidata\_tekgen/ontologies}}.

\vspace{-20pt}

\paragraph{\textbf{Triple generation and alignment with sentences}}

Given an ontology from the previous step, a parameterized SPARQL query\footnote{\url{https://github.com/cenguix/Text2KGBench/tree/main/src/benchmark}} is used to generate a set of K triples for each of the relations. The SPARQL query guaranteed that the triples confirmed the domain and range restrictions of each ontology. For example, for ``director'' relation, we would get triples such as director(``Lion King'',``Roger Allers'').

In this dataset, we reused the TekGen corpus \cite{agarwal-etal-2021-knowledge} which provides Wikidata triples aligned with corresponding sentences from Wikipedia. The TekGen corpus is generated using distant supervision and it has 16 M aligned triple-sentences covering 663 Wikidata relations.  For each triple we got from the previous step, we analyzed the TekGen corpus to get an aligned sentence when available. For instance, the triple in the previous sentence will be aligned to a sentence such as ``The Lion King is an animated musical drama film directed by Roger Allers and Rob Minkoff, produced by Don Hahn.``. Once a sentence is found, we check all the other relations associated with the sentence in the TekGen corpus and include them also if they are part of our ontology. For example, in this sentence, director (``Lion King'', ``Rob Minkoff'') and producer(``Lion King'', ``Don Hahn'') will also be included in the dataset. 

Once we complete this process for all 10 ontologies, we generated 13,474 sentence - triple(s) alignments and they are divided into train, validation and test sets.

\paragraph{\textbf{Manual Validations and cleaning}}

Because the TekGen corpus is generated using distant supervision, it can have noise and some incorrect alignments. In order to evaluate models with a more precise set of test cases, we have manually analyzed the test sentences and selected a smaller subset of more accurately aligned sentences for each ontology. For this exercise, the annotators looked at the triple and aligned sentence in the gold standard and selected sentences that a human can easily extract the triple such that the fact is explicitly mentioned in the text. For example, ``The film was also nominated for Academy Award for Best Picture.'' is a noisy sentence to extract the triple ``nominated for(Working Girl, Academy Award for Best Picture) as it is impossible for a model to resolve coreference to understand what term ``the film'' is referring to, only with this sentence as input. Another example, the sentence ``Welcome to Eltingville was written by Dorkin and Chuck Sheetz'' is wrongly aligned with the triple director(``Welcome to Eltingville'', ``Chuck Sheetz'') because the entities co-occur in the sentence and Chuck Sheetz is both the director and the writer. For a sample of test data, the authors removed such alignments and created another test set with 939 verified sentence-triple alignments.  The systems can use both the larger test set and this smaller high-quality test set for their evaluations. 

\paragraph{\textbf{Unseen sentence generation}}

One of the caveats of this benchmark is that the language models under test might have already seen these sentences or even the alignments in some form. Then it can be argued that they might have memorized some of these relations. One important aspect to evaluate is if the model performance will get affected if we test the model with unseen sentences that are not part of Wikipedia and not seen during the pre-training. For that, we invent new sentences with facts that the annotators come up with. For example, a sentence such as ``John Doe starred in the movie The Fake Movie released in 2025''. With this exercise, the authors generated 174 unseen sentences roughly two sentences per each relation in each ontology. Furthermore, this unseen set of sentences can be used to check how faithful the model is to the given sentence when generating the triples.

\begin{table}[]
\caption{Statistics related to the two datasets including the list of ontologies, number of types and relations in each ontology, and number of sentences aligned.}
\label{tab:onto_stats}
\begin{tabular}{|lccc|lccclccc|}
\hline
\multicolumn{4}{|c|}{wikidata-tekgen} & \multicolumn{8}{c|}{dbpedia-webnlg} \\ \hline
\multicolumn{1}{|l|}{Ontology} & \multicolumn{1}{c|}{Types} & \multicolumn{1}{c|}{Rels.} & Sents. & \multicolumn{1}{l|}{Ontology} & \multicolumn{1}{c|}{Types} & \multicolumn{1}{c|}{Rels} & \multicolumn{1}{c|}{Sents.} & \multicolumn{1}{l|}{Ontology} & \multicolumn{1}{c|}{Types} & \multicolumn{1}{c|}{Rels} & Sents. \\ \hline
\multicolumn{1}{|l|}{Movie} & \multicolumn{1}{c|}{12} & \multicolumn{1}{c|}{15} & 2800 & \multicolumn{1}{l|}{University} & \multicolumn{1}{c|}{15} & \multicolumn{1}{c|}{46} & \multicolumn{1}{c|}{156} & \multicolumn{1}{l|}{Transport} & \multicolumn{1}{c|}{20} & \multicolumn{1}{c|}{68} & 314 \\ \hline
\multicolumn{1}{|l|}{Music} & \multicolumn{1}{c|}{13} & \multicolumn{1}{c|}{13} & 2243 & \multicolumn{1}{l|}{Music} & \multicolumn{1}{c|}{15} & \multicolumn{1}{c|}{35} & \multicolumn{1}{c|}{290} & \multicolumn{1}{l|}{Monument} & \multicolumn{1}{c|}{14} & \multicolumn{1}{c|}{26} & 92 \\ \hline
\multicolumn{1}{|l|}{Sport} & \multicolumn{1}{c|}{15} & \multicolumn{1}{c|}{11} & 1693 & \multicolumn{1}{l|}{Airport} & \multicolumn{1}{c|}{14} & \multicolumn{1}{c|}{39} & \multicolumn{1}{c|}{306} & \multicolumn{1}{l|}{Food} & \multicolumn{1}{c|}{12} & \multicolumn{1}{c|}{24} & 398 \\ \hline
\multicolumn{1}{|l|}{Book} & \multicolumn{1}{c|}{20} & \multicolumn{1}{c|}{12} & 1810 & \multicolumn{1}{l|}{Building} & \multicolumn{1}{c|}{14} & \multicolumn{1}{c|}{38} & \multicolumn{1}{c|}{275} & \multicolumn{1}{l|}{Written Work} & \multicolumn{1}{c|}{10} & \multicolumn{1}{c|}{44} & 322 \\ \hline
\multicolumn{1}{|l|}{Military} & \multicolumn{1}{c|}{13} & \multicolumn{1}{c|}{9} & 750 & \multicolumn{1}{l|}{Athlete} & \multicolumn{1}{c|}{17} & \multicolumn{1}{c|}{37} & \multicolumn{1}{c|}{293} & \multicolumn{1}{l|}{Sports Team} & \multicolumn{1}{c|}{14} & \multicolumn{1}{c|}{24} & 235 \\ \hline
\multicolumn{1}{|l|}{Computer} & \multicolumn{1}{c|}{15} & \multicolumn{1}{c|}{12} & 743 & \multicolumn{1}{l|}{Politician} & \multicolumn{1}{c|}{19} & \multicolumn{1}{c|}{40} & \multicolumn{1}{c|}{319} & \multicolumn{1}{l|}{City} & \multicolumn{1}{c|}{11} & \multicolumn{1}{c|}{23} & 348 \\ \hline
\multicolumn{1}{|l|}{Space} & \multicolumn{1}{c|}{15} & \multicolumn{1}{c|}{7} & 666 & \multicolumn{1}{l|}{Company} & \multicolumn{1}{c|}{10} & \multicolumn{1}{c|}{28} & \multicolumn{1}{c|}{153} & \multicolumn{1}{l|}{Artist} & \multicolumn{1}{c|}{20} & \multicolumn{1}{c|}{39} & 386 \\ \hline
\multicolumn{1}{|l|}{Politics} & \multicolumn{1}{c|}{13} & \multicolumn{1}{c|}{9} & 695 & \multicolumn{1}{l|}{Celestial} & \multicolumn{1}{c|}{8} & \multicolumn{1}{c|}{27} & \multicolumn{1}{c|}{194} & \multicolumn{1}{l|}{Scientist} & \multicolumn{1}{c|}{15} & \multicolumn{1}{c|}{47} & 259 \\ \hline
\multicolumn{1}{|l|}{Nature} & \multicolumn{1}{c|}{14} & \multicolumn{1}{c|}{13} & 1558 & \multicolumn{1}{l|}{Astronaut} & \multicolumn{1}{c|}{16} & \multicolumn{1}{c|}{38} & \multicolumn{1}{c|}{154} & \multicolumn{1}{l|}{Film} & \multicolumn{1}{c|}{18} & \multicolumn{1}{c|}{44} & 264 \\ \hline
\multicolumn{1}{|l|}{Culture} & \multicolumn{1}{c|}{15} & \multicolumn{1}{c|}{8} & \multicolumn{1}{c|}{516} & \multicolumn{1}{l|}{Comics} & \multicolumn{1}{c|}{10} & \multicolumn{1}{c|}{18} & \multicolumn{1}{c|}{102} & \multicolumn{1}{l|}{} & \multicolumn{1}{l|}{} & \multicolumn{1}{l|}{} & \multicolumn{1}{l|}{} \\ \hline
\multicolumn{3}{|l|}{Total} & \multicolumn{1}{l|}{13,474} & \multicolumn{7}{l|}{Total} & \multicolumn{1}{l|}{4,860} \\ \hline
\end{tabular}
\label{tab:onto_stats}
\end{table}

\subsection{DBpedia-WebNLG Dataset}

The DBpedia-WebNLG dataset is created reusing the alignments in the WebNLG corpus.

\paragraph{\textbf{Ontology Selection}} Similar to the previous dataset, the first step is to create a set of ontologies. WebNLG consists of 19 categories and we created an ontology for each category. First, we analysed the triples in each category to extract the relations in each category and defined the concepts based on the domain and range constraints of those relations. The statistics for the resulting 19 ontologies are shown in Table~\ref{tab:onto_stats}. 

\paragraph{\textbf{Triple generation and alignment with sentences}} We have parsed the WebNLG 3.0 English dataset and collected the sentences in one of the splits (WebNLG 3triples). When creating train and test sets, we made sure that the same fact would not appear in both train and test sets. Because the alignments (verbalizations) are verified by crowdsourcing in WebNLG, there was no need for us to create a manually validated set. We generated 4,860 sentence - triple(s) alignments using WebNLG data and divided into train and test splits. 

The train/val/test splits for both benchmarks were done as stratified randomized folds aiming to preserve the relation distributions as much as possible using scikit-learn. The rationale for the splits was to provide training data (examples for in-context learning or fine-tuning models) for future systems that will use the benchmark and validation data (for optimizing hyperparameters).

\section {Evaluation Metrics}
\label{sec.eval.metrics}
In this section, we present the set of evaluation metrics we use in \textit{Text2KGBench} to measure the performance of systems for generating facts from the text. Evaluation metrics aim to validate three aspects: (i) extracted facts are accurate according to the given ontology, (ii) extracted facts conform to the given ontology, and (iii) the output doesn't include any hallucinations. 

Given a prompt similar to Figure~\ref{fig:prompt}, the LLM will produce a textual output that can be parsed into a set of triples, which we call LLM output triples. The expected output for each test sentence is in the ground truth files.

\paragraph{\textbf{Fact Extraction Accuracy:}} This is measured using Precision (P), Recall (R), and F1 scores by comparing LLM output triples to the ground truth triples. P is calculated by dividing the number of correct LLM triples (which are part of the ground truth) by the number of LLM triples. R is calculated by dividing the number of correct LLM triples by the number of ground truth triples. F1 is calculated as the harmonic mean of the P and R. If the LLM output is empty, P, R, and F1 are set to 0. Because the set of triples are not exhaustive for a given sentence, to avoid false negatives, we follow a locally closed approach by only considering the relations that are part of the ground truth. For P, R, F1, higher numbers represent better performance. 

\paragraph{\textbf{Ontology Conformance:}} This is measured using the Ontology Conformance (OC) metric which is calculated as the percentage of LLM output triples conforming to the input ontology, i.e., ontology conforming LLM output triples divided by total LLM output triples. In this version, a triple is considered to be conforming to the ontology if the relation is one of the canonical relations listed in the ontology. This can be further extended to validate other restrictions such as domain, range or other ontological axioms. 

\paragraph{\textbf{Hallucinations:}} Hallucination is defined as the generated content that is nonsensical or unfaithful to the provided source content \citep{10.1145/3571730}. We calculate three hallucination metrics, subject hallucination (SH), relation hallucination (RH), and object hallucination (OH). These are calculated by comparing the generated triple to the test sentence and the ontology. For each triple, SH and OH check if the subject and object are present in either the sentence or the ontology concepts, and RH checks if the relation is present in the ontology relations. For SH and OH, we use stemming to account for inflected forms with morphological variations such as ``America'', ``American'', ``Americans'', etc. Each term in the subject or object and test sentence is stemmed before checking if the subject or object is present as a substring in the test sentence and/or ontology concepts. For RH, relations are matched using exact matches. In this version, RH and OC are inversely related \textit{i.e.} 1 - OC equals to RH.

\section{Baselines and Evaluation Results}
\label{sec.eval.results}

% This section is critical where we obtain several important figures spanning from novel ones such as Ontology Conformance, Ontology Relation Hallucination, Ontology Subject/Object Triple Hallucination and the typical Precision, Recall and F1 score measurements. The whole pipeline stages involved in calculating such figures include the chosen text corpus namely the adapted and post-processed TekGen corpus, the dynamically generated automatic prompt generation, the Vicuna LLM model chosen obtained via the model incremental weights to the Llama LLM. 
% Finally as expected, in this section we present the entire set of figures obtained mentioned above.

In this section, we present baseline LLM models, baselines for automatic prompt generation and the evaluation results for baseline models for the two datasets of \textit{Text2KGBench} we described in Section~\ref{sec.bench.gen}.

\subsection{Baseline LLM Models}

\subsubsection{Vicuna-13B} Vicuna-13B \citep{vicuna2023} is an open-source LLM that fine-tunes a base LLaMA model with 70K user-shared conversations from ShareGPT. We obtained the LlaMA 13B LLM model, checkpoints, and tokenizer, through the Pyllama Github repository\footnote{\url{https://github.com/juncongmoo/pyllama}} and applied Vicuna weights from FastChat~\footnote{\url{https://github.com/lm-sys/FastChat}} as delta weights. Vicuna-13B claims 90\% performance of OpenAI ChatGPT and Google Bard \citep{zheng2023judging} where  the authors have used a metric “Relative Response Quality" using strong LLM (GPT4) as judges to evaluate the model on open-ended questions.

%The model used in this paper is the Vicuna-13B~\cite{vicuna2023} an open-source LLM trained by fine-tuning LLaMA on 70K conversations from user-shared ChatGPT conversations. A preliminary evaluation demonstrated that Vicuna-13B achieves about 90\% performance of OpenAI ChatGPT and Google Bard and outperforms both LLaMA and Stanford Alpaca LLMs. 
% Vicuna LLM model training is based upon Stanford’s Alpaca LLM by improving memory optimizations enabling the maximum context length from 512 to 2048, by allowing multi-round conversations and reducing training costs.

\subsubsection{Alpaca-LoRA-13B} Alpaca-LoRA~\footnote{\url{https://github.com/tloen/alpaca-lora}} is a model that fine-tuned a base LLaMA model with the same 52K instructions of Alpaca model that is generated using self-instruct \citep{wang2022self} with the OpenAI's \textit{text-davinci-003} model. Alpaca-LoRA is fine-tuned using Low-Rank Adaptation \citep{hu2022lora} allows reducing the number of trainable parameters by a significant order by freezing the pre-trained model weights and injecting trainable rank decomposition matrices to each transformer layer. 

\subsection{Automatic Prompt Generation}

Both our LLM models are GPT-style decoder-only models that are instruction fine-tuned. They can be used for downstream tasks by providing a prompt with an instruction. In this section, we present the steps involved in automatically creating prompts for each test sentence. 

Our baseline prompt consists of our main parts: (a) Instruction, (b) Ontology description, (c) Demonstrative examples, and (d) Test sentence as illustrated in Figure~\ref{fig:prompt}.

\paragraph{\textbf{Instruction}} This is a fixed instruction that we used for all test cases across the ontologies. We used the following phrase ``Given the following ontology and sentences, please extract the triples from the sentence according to the relations in the ontology. In the output, only include the triples in the given output format.'' as the instruction. We describe the task as well as request the model to be less verbose and output only the triples in the given format. 

\paragraph{\textbf{Ontology description}} This part of the prompt provides a description of the ontology to the model as context. Each test case in our benchmark is associated with an ontology. This part of the prompts verbalizes the ontology by listing the set of concepts, and a set of relations with their domain and range constraints given by the ontology. For example, for the test case in the movie ontology, concepts will be a list such as a film, film genre, genre, film production company, film award, human etc. and the relations will be a list such as director(film, human), cast\_member(film, human), award\_received(film, award), genre(film, genre), production\_company(film, film production company), etc. Throughout the prompt, we use the \textit{relation(subject, object)} notation for representing relations and expect the model to follow the notation in the output.

\paragraph{\textbf{Demonstrative Examples}} This part of the prompt is used to provide the LLM with an example to show an input sentence and the expected output. LLMs are capable of In-Context Learning where they learn the task and output format from the examples provided in the prompt. The examples are taken from the training data for each of the datasets based on their similarity to the test sentence. We have used sentence similarities using Sentence Transformers (SBERT) \citep{reimers-2019-sentence-bert} with the T5-XXL model \citep{ni-etal-2022-large}. For instance, given a test sentence such as ``Super Capers, a film written by Ray Griggs, is 98 minutes long.'', it can find the most similar sentence in training data, ``English Without Tears, written by Terence Rattigan, runs 89 minutes.'' with it's aligned triples. Example output follows the same relation notation.  

\paragraph{\bf{Test Sentence}} Finally, the prompt contains the test sentence from which we want to extract the facts complying with the ontology. Similar to the example, the prompt ends with a ``Test Output:`` where the model is expected to generate the facts in the sentence following the same format as in the example sentence. 

\subsection{Evaluation Results}
\label{sec.results}

We run inferences for automatically generated prompts for both \textit{Wikidata-TekGen} and \textit{DBpedia-WebNLG} corpora and calculate the metrics discussed in Section~\ref{sec.eval.metrics}: Precision (P), Recall (R), F1, Ontology Conformance (OC), Subject/Relation/ Object Hallucinations (SH/RH/OH). Table~\ref{tab:llm-total-avg} illustrates the average values across all ontologies in a given dataset. As discussed in Section~\ref{sec.bench.gen}, three different settings in \textit{Wikidata-TekGen} dataset: all test cases (All), manually validated and cleaned subset (selected), and unseen sentences (Unseen) which annotated in the ``Variant'' column. 

Each row in Table~\ref{tab:llm-total-avg} is an aggregation of results from test cases across multiple ontologies (10 for Wikidata-TekGen and 19 for DBpedia-WebNLG) and Table~\ref{tab:vicuna-all-tests} shows the results at each individual ontology level for the first row of Table~\ref{tab:llm-total-avg}, \textit{i.e.}, Wikidata-TekGen - Vicuna - All. For brevity, ontology-level results for other rows are included in the project Wiki\footnote{\url{https://github.com/cenguix/Text2KGBench/wiki}}.

\begin{table}[!hbt]
	\centering
	\caption{This table summarizes average evaluation metrics for all ontologies in Wikidata-TekGen and the DBpedia-WebNLG datasets.}
	\label{tab:llm-total-avg}
	\begin{tabular}{|c|lc|ccc|c|ccc|}
		\hline
		\multirow{2}{*}{\textbf{Dataset}} & \multicolumn{1}{l|}{\multirow{2}{*}{\textbf{Model}}} & \multicolumn{1}{l|}{\multirow{2}{*}{\textbf{Variant\tablefootnote{Refer to Section~\ref{sec.bench.gen} for details.}}}} & \multicolumn{3}{c|}{\textbf{Fact Extraction}} & \multirow{2}{*}{\textbf{\begin{tabular}[c]{@{}c@{}}OC\end{tabular}}} & \multicolumn{3}{c|}{\textbf{Hallucinations}} \\ \cline{4-6} \cline{8-10} 
		& \multicolumn{1}{l|}{} & \multicolumn{1}{l|}{} & \multicolumn{1}{c|}{\textbf{P}} & \multicolumn{1}{c|}{\textbf{R}} & \textbf{F1} &  & \multicolumn{1}{c|}{\textbf{SH}} & \multicolumn{1}{c|}{\textbf{RH}} & \textbf{OH} \\ \hline
		\multirow{6}{*}{\textbf{Wikidata-TekGen}} & \multicolumn{1}{l|}{\multirow{3}{*}{\textbf{Vicuna}}} & \textbf{All} & \multicolumn{1}{c|}{0.38} & \multicolumn{1}{c|}{0.34} & 0.35 & 0.83 & \multicolumn{1}{c|}{0.17} & \multicolumn{1}{c|}{0.17} & 0.17 \\ \cline{3-10} 
		& \multicolumn{1}{l|}{} & \textbf{Selected} & \multicolumn{1}{c|}{0.42} & \multicolumn{1}{c|}{0.39} & 0.38 & 0.84 & \multicolumn{1}{c|}{0.11} & \multicolumn{1}{c|}{0.16} & 0.14 \\ \cline{3-10} 
		& \multicolumn{1}{l|}{} & \textbf{Unseen} & \multicolumn{1}{c|}{0.32} & \multicolumn{1}{c|}{0.32} & 0.32  & 0.86 & \multicolumn{1}{c|}{0.07} & \multicolumn{1}{c|}{0.14} & 0.14  \\ \cline{2-10} 
		& \multicolumn{1}{l|}{\multirow{3}{*}{\textbf{\begin{tabular}[c]{@{}c@{}}Alapaca\\LoRA\end{tabular}}}} & \textbf{All} & \multicolumn{1}{c|}{0.32} & \multicolumn{1}{c|}{0.26} & 0.27 & 0.87 & \multicolumn{1}{c|}{0.18} & \multicolumn{1}{c|}{0.13} & 0.17 \\ \cline{3-10} 
		& \multicolumn{1}{l|}{} & \textbf{Selected} & \multicolumn{1}{c|}{0.33} & \multicolumn{1}{c|}{0.27} & 0.28 & 0.87 & \multicolumn{1}{c|}{0.12} & \multicolumn{1}{c|}{0.13} & 0.17 \\ \cline{3-10} 
		& \multicolumn{1}{l|}{} & \textbf{Unseen} & \multicolumn{1}{c|}{0.22} & \multicolumn{1}{c|}{0.22} & 0.22  & 0.86 & \multicolumn{1}{c|}{0.09} & \multicolumn{1}{c|}{0.14} & 0.26 \\ \hline
		\multirow{2}{*}{\textbf{DBpedia-WebNLG}} & \multicolumn{2}{c|}{\textbf{Vicuna}} & \multicolumn{1}{c|}{0.34} & \multicolumn{1}{c|}{0.27} & 0.30 & 0.93 & \multicolumn{1}{c|}{0.12} & \multicolumn{1}{c|}{0.07} & 0.28  \\ \cline{2-10} 
		& \multicolumn{2}{c|}{\textbf{Alpaca-LoRA}} & \multicolumn{1}{c|}{0.32} & \multicolumn{1}{c|}{0.23} & 0.25 & 0.91 & \multicolumn{1}{c|}{0.16} & \multicolumn{1}{c|}{0.09} & 0.38 \\ \hline
	\end{tabular}
\end{table}

\begin{table}[]
	\centering
	\caption{Results for Vicuna LLM All Test Cases. Numbers in bold identify the best results for each metric. Numbers underlined identify worst results.}
	\label{tab:vicuna-all-tests}
	\begin{tabular}{|l|ccccccc}
		\hline
		\multicolumn{1}{|c|}{\multirow{2}{*}{Ontology}} & \multicolumn{3}{l|}{Fact Extraction} & \multicolumn{1}{l|}{\multirow{2}{*}{\begin{tabular}[c]{@{}l@{}}OC\end{tabular}}} & \multicolumn{3}{l|}{Hallucinations} \\ \cline{2-4} \cline{6-8} 
		\multicolumn{1}{|c|}{} & \multicolumn{1}{c|}{P} & \multicolumn{1}{c|}{R} & \multicolumn{1}{c|}{F1} & \multicolumn{1}{l|}{} & \multicolumn{1}{c|}{SH} & \multicolumn{1}{c|}{RH} & \multicolumn{1}{c|}{OH} \\ \hline
		\multicolumn{1}{|l|}{1. Movie Ontology} & \multicolumn{1}{c|}{0.33} & \multicolumn{1}{c|}{\underline{0.23}} & \multicolumn{1}{c|}{0.25} & \multicolumn{1}{c|}{0.89} & \multicolumn{1}{c|}{\underline{0.26}} & \multicolumn{1}{c|}{0.11} & \multicolumn{1}{c|}{\underline{0.26}} \\ \hline
		\multicolumn{1}{|l|}{2. Music Ontology} & \multicolumn{1}{c|}{0.42} & \multicolumn{1}{c|}{0.28} & \multicolumn{1}{c|}{0.32} & \multicolumn{1}{c|}{\textbf{0.94}} & \multicolumn{1}{c|}{0.16} & \multicolumn{1}{c|}{\textbf{0.06}} & \multicolumn{1}{c|}{0.22} \\ \hline
		\multicolumn{1}{|l|}{3. Sport Ontology} & \multicolumn{1}{c|}{0.57} & \multicolumn{1}{c|}{0.52} & \multicolumn{1}{c|}{0.52} & \multicolumn{1}{c|}{0.85} & \multicolumn{1}{c|}{0.22} & \multicolumn{1}{c|}{0.15} & \multicolumn{1}{c|}{0.13} \\ \hline
		\multicolumn{1}{|l|}{4. Book Ontology} & \multicolumn{1}{c|}{0.31} & \multicolumn{1}{c|}{0.25} & \multicolumn{1}{c|}{0.26} & \multicolumn{1}{c|}{0.92} & \multicolumn{1}{c|}{0.16} & \multicolumn{1}{c|}{0.08} & \multicolumn{1}{c|}{0.23} \\ \hline
		\multicolumn{1}{|l|}{5. Military Ontology} & \multicolumn{1}{c|}{\underline{0.24}} & \multicolumn{1}{c|}{0.25} & \multicolumn{1}{c|}{\underline{0.24}} & \multicolumn{1}{c|}{0.8} & \multicolumn{1}{c|}{0.19} & \multicolumn{1}{c|}{0.2} & \multicolumn{1}{c|}{\underline{0.26}} \\ \hline
		\multicolumn{1}{|l|}{6. Computer Ontology} & \multicolumn{1}{c|}{0.38} & \multicolumn{1}{c|}{0.35} & \multicolumn{1}{c|}{0.35} & \multicolumn{1}{c|}{0.85} & \multicolumn{1}{c|}{0.15} & \multicolumn{1}{c|}{0.15} & \multicolumn{1}{c|}{0.11} \\ \hline
		\multicolumn{1}{|l|}{7. Space Ontology} & \multicolumn{1}{c|}{\textbf{0.68}} & \multicolumn{1}{c|}{\textbf{0.67}} & \multicolumn{1}{c|}{\textbf{0.66}} & \multicolumn{1}{c|}{0.93} & \multicolumn{1}{c|}{0.15} & \multicolumn{1}{c|}{0.07} & \multicolumn{1}{c|}{\textbf{0.08}} \\ \hline
		\multicolumn{1}{|l|}{8. Politics Ontology} & \multicolumn{1}{c|}{0.34} & \multicolumn{1}{c|}{0.32} & \multicolumn{1}{c|}{0.33} & \multicolumn{1}{c|}{0.92} & \multicolumn{1}{c|}{0.17} & \multicolumn{1}{c|}{0.08} & \multicolumn{1}{c|}{0.15} \\ \hline
		\multicolumn{1}{|l|}{9. Nature Ontology} & \multicolumn{1}{c|}{0.25} & \multicolumn{1}{c|}{0.27} & \multicolumn{1}{c|}{0.25} & \multicolumn{1}{c|}{0.68} & \multicolumn{1}{c|}{\textbf{0.1}} & \multicolumn{1}{c|}{0.32} & \multicolumn{1}{c|}{0.14} \\ \hline
		\multicolumn{1}{|l|}{10 Culture Ontology} & \multicolumn{1}{c|}{0.31} & \multicolumn{1}{c|}{0.32} & \multicolumn{1}{c|}{0.31} & \multicolumn{1}{c|}{\underline{0.59}} & \multicolumn{1}{c|}{0.15} & \multicolumn{1}{c|}{\underline{0.41}} & \multicolumn{1}{c|}{0.12} \\ \hline
		\multicolumn{1}{|l|}{Ontologies Average} & \multicolumn{1}{c|}{0.38} & \multicolumn{1}{c|}{0.34} & \multicolumn{1}{c|}{0.35} & \multicolumn{1}{c|}{0.83} & \multicolumn{1}{c|}{0.17} & \multicolumn{1}{c|}{0.17} & \multicolumn{1}{c|}{0.17} \\ \hline
	\end{tabular}
\end{table}

From the results, some initial observations from Table~\ref{tab:llm-total-avg} on the different datasets and LLM models: 
\begin{itemize}
	\item{Precision, Recall and F1 score have low intermediate values}
	\item{Ontology Conformance is pretty high in almost all entries}
	\item{Subject, Relation, Object Hallucination is relatively low}
\end{itemize}

These results should be further analyzed to understand the different capbalities and limitations of LLMs in KG generation from text. An in-depth analysis of the results is out of scope of this paper due to space limitations and we expect the system papers using the benchmark to provide insights and conclusions on this aspect. As we have used LLM models as is without any fine-tuning, prompt tuning or semantic validation, we believe there is a large room for improvements.

% In the different LLM models, we observe that the LLM triples returned by the LLM include some mistakes, in spite of the fact that the Ontology Conformance is pretty high in almost all ontologies and LLM models tested. 
% In other words, we derive that subject/relation/object hallucinations negatively affect the Precision, Recall, and F1 score of LLMs, even considering that hallucination figures are relatively low. 
  
\subsection{Error Analysis}
We have performed an initial error analysis to understand the errors made by the models and Table~\ref{tab:errors} shows some examples for different types of errors. In addition, we noticed that there are some false positives in hallucination due to LLMs expanding acronyms, for example, the sentence can have ``NATO'' where the model generates ``North Atlantic Treaty Organization'' as a subject. We plan to consider acronyms, aliases, etc. in hallucination calculations in the future.

\begin{table}[h]
\caption{Examples of errors from the Vicuna13B model with Wikidata-TekGen}
\begin{tabular}{|p{6cm}|p{4.5cm}|p{4.5cm}|}
\hline
\small Sentence & \small Triple & \small Error Type \\ \hline
 Aparajito won 11 international awards, including the Golden Lion and Critics Award at the Venice Film Festival, becoming the first ever film to win both. &  award\_received(Aparajito, Venice Film Festival) &  An incorrect fact extracted. The model mistook the film festival for an award. \\ \hline
 The Gallopin Gaucho was a second attempt at success by co-directors Walt Disney and Ub Iwerks. &  directed(The Gallopin Gaucho,Walt Disney) &  Ontology conformance error. The canonical relation is the director. \\ \hline
 American Born Chinese is a graphic novel by Gene Luen Yang. &  narrative\_location(American Born Chinese, San Francisco) &  Object hallucination. Neither the object nor the relation is mentioned in the text. \\ \hline
 Schreck was a founding member of the Sturmabteilung. & member\_of\_political\_party (Hermann Goring, Sturmabteilung) & Subject hallucination. Hermann Goring is not mentioned in the text.  \\ \hline
\end{tabular}
\label{tab:errors}
\end{table}

\section {Related Work}
\label{sec.related.work}
The primary aim of the knowledge graph generation task is to extract structured information from heterogeneous sources. This section will explore the Relation Extraction Benchmarks, Foundation Models for Knowledge Graph Generation and Semi-Automatic/Automatic Knowledge Graph Completion (KBC)-KG-triple generation. 
Relation extraction has made substantial use of the following datasets such as  the New York Times (NYT)/ NYT-FB dataset  \citep{mintz2009distant} \citep{riedel2010modeling} \citep{marcheggiani2016discrete}, TAC Relation Extraction Dataset (TACRED) \citep{zhang2017position}, Large-Scale Document-Level Relation Extraction Dataset(DocRED) \citep{yao2019docred}, The WEB-NLG dataset \citep{gardent2017creating}, FewRel dataset \citep{han2018fewrel}, FewRel 2.0 \citep{gao2019fewrel}. The relation extraction benchmarks that exist for the scientific domain are SciERC dataset \citep{luan2018multi} and SCIREX \citep{jain2020scirex}. The SCIREX dataset is intended to detect both binary and n-ary relations between entities and concepts, while the SciERC dataset is intended to identify binary relations between entities in scientific papers.  There are few datasets that cover multiple languages, such as Multilingual LAMA (Language Model Analysis) dataset \citep{kassner2021multilingual}, which cover 53 languages, MiLER SMiLER(Samsung Multi-Lingual Entity and Relation Extraction dataset \citep{seganti2021multilingual} covers 14 languages, DiS-ReX \citep{bhartiya2022dis} covers 4 languages.
 Through entity linking, Knowledge-Enhanced Relation Extraction Dataset (KERED) \citep{lin2022knowledge} gives knowledge context for entities and annotates each sentence with a relational fact. This dataset consists of NYT10m, Wikidata \citep{vrandevcic2014wikidata} (Wiki80 and Wiki20m).

Relation extraction benchmark datasets can be used to evaluate the performance of foundation models. A survey paper \citep{yang2023harnessing} has explored the history of these foundation models and summarizes different tasks. Foundation models are generally categorized into two categories: Encoder-only or Encoder-Decoder (BERT style) and Decoder-only (GPT style) \citep{yang2023harnessing}. BERT-style models are still challenging as they are under development and mostly available as open-source. They are considered as Mask Language Models that include RoBERTa \citep{liu2019roberta}, BERT \citep{devlin2018bert}, and T5  \citep{raffel2020exploring}. Decoder-only (GPT style) models (GPT-3 \citep{brown2020language},  PaLM \citep{chowdhery2022palm}, OPT \citep{zhang2022opt} and BLOOM \citep{scao2022bloom}) generally need finetuning on datasets of the particular downstream task. Brown et. al. \citep{brown2020language} have trained GPT-3 (an autoregressive language model) with 175 billion parameters and also tested its performance with the few-shot setting. Jeremy and Sebastian \citep{howard2018universal} have proposed an effective transfer learning method, Universal Language Model Fine-tuning (ULMFiT), for any NLP task. Brian et. al. \citep{lester2021power} explores the prompt tuning to learn soft prompts for adapting language models. Soft prompts are learned by back propagation, while GPT-3 uses discrete text prompts \citep{wang2019superglue}.\\
Vicuna \citep{vicuna2023} is an open-source chatbot and it is trained by fine-tuning LLaMA. It is shown by evaluation that Vicuna has performed more than 90\% quality of Google Bard  and OpenAI ChatGPT compared to other models like LLaMA and Alpaca. Alpaca \citep{alpaca} has been introduced as a strong, replicable instruction-following model. It is fine-tuned from the LLaMA 7B model on 52K instruction-following demonstrations.

KBC and KG-triple generation have become a hot research field with the synergy/integration with LLMs. The possibilities are limitless regarding the automatic generation of new triples via the use of LLMs and the only \textit{"Achilles Heel"} consists of computer resources required for integrating both systems. In \citep{bi2023codekgc} is presented a system that generates automatically triples from natural language and code completion tasks. In this case, it is presented as input code excerpts denoting class and function definitions. They consider the use of neural networks present in pre-trained LLM's as \textit{"black boxes"}. In \citep{alivanistos2022prompting} is presented a system that uses a GPT3 LLM with the aim of building a knowledge base semi-automatically via a multi-step process combining customized prompting techniques for predicting missing objects in triples where subjects and relations are given. In~\citep{Khorashadizadeh2023}, the authors perform a qualitative study of large language models using ChatGPT for various tasks including KG population, KG completion, triple or fact verification and identify some challenges such as hallucination, fairness and bias, and high computational cost. And finally, we include reference \citep{veseli2023evaluating} in this section where it is presented a benchmark dataset for assessing Knowledge Base Completion (KBC) potential for language models (LM).

Regarding hallucination metrics, Deep Learning based system-generated data is sensitive to hallucinate \textit{“unfaithful”} text degrading system performance, especially in critical systems such as patient data where untruthful data could pose serious risks to patients. 
In \cite{10.1145/3571730} it is presented a survey describing hallucination metrics and generic downstream Natural Language Generation (NLG) tasks sensitive to the hallucination phenomena, such as Generative Question Answering (GQA), Dialogue Generation, Data-to-text Generation, and so on. The authors make evidence of the difficulty that poses detecting nonsensical/unfaithful texts to source content in NLG/Transformer-based systems. The authors make a distinction between \textit{"intrinsic"} and \textit{"extrinsic"} hallucinations, the former output generator not \textit{“faithful”} to the source and the latter output that cannot be confirmed, nor negated from the source content referring to for instance real-world external factual sources. In our case, we focus on intrinsic hallucinations and include in the future work section extrinsic hallucination detection. We do not make a distinction as provided in the survey, between \textit{“faithfulness”} and \textit{“factuality”}. We use the source input/ground truth as the fact (i.e. for instance ontology conformance) indicating the degree of \textit{“faithfulness”} of such data. Regarding the method for the hallucination metrics measurements, our system is based upon model-based metrics such as Information Extraction-based representing knowledge as Semantic Web triple tuples (i.e., for instance: relation(subject, object) triples aided by stemming pre-processing). We also foresee more fuzzy-based faithfulness/hallucination metrics by applying vector-based knowledge embeddings and retrieving via K-Nearest-Neighbors (KNN) the most related triples with a given threshold.

\section{Conclusion and Future Work}
\label{sec.conclusions}

In this paper, we presented \textit{Text2KGBench}, a benchmark for evaluating capabilities of LLMs for extracting facts from a text corpora guided by an ontology.

\paragraph{Limitations} In this version, we have only considered smaller-sized ontologies by design to cater for the token size limitations of LLMs. Nevertheless, in practice, there are quite larger ontologies in domains such as medicine. In future versions, we plan to include cases with much larger ontologies which will require systems to automatically select the portion of the ontology or the set of axioms that are relevant to the given input text. In addition, there is research on extending the capabilities of LLMs to handle longer contexts such as Unlimiformer\citep{bertsch2023unlimiformer}. Furthermore, in this version, we have separated the OWL/RDF representations of KGs by verbalizing ontologies and triples. In future versions, we will test LLMs on handing these representations directly without pre/post-processing.   

\paragraph{Future Work}
One important aspect when it comes to foundation models is bias and fairness. In future work, we would like to further extend our benchmark considering different bias variables such as gender, race/ethnicity, geographic location, etc. and create contrastive test cases to verify the fairness of LLMs when generating Knowledge Graphs from text. In other words, we would like to systematically evaluate if this process performs better for a certain subgroup based on their  gender, demographics, or socioeconomic status. Furthermore, we will plan to measure more reasoning capabilities when performing fact extraction and KG generation. We plan to extend the benchmark with a dataset that requires more semantic reasoning to perform the task. In the Text2KGBench benchmark we have currently focused on available open-source LLM models. In addition, we plan to compare both LLM base-lines, Vicuna-13B and Alpaca-Lora-13B, and any emerging new open-source LLMs to the commercial OpenAI's ChatGPT\footnote{\url{https://openai.com/blog/chatgpt}} and GPT-4\footnote{\url{https://openai.com/gpt-4}} LLMs.

\paragraph*{Impact:} With the popularity of GPT-like LLMs, there is a big enthusiasm for using such models jointly with KGs and for constructing KGs. Authors firmly believe that ontology-driven KG construction from text leveraging LLMs will be of interest to the Semantic Web community. To the best of our knowledge, \textit{Text2KG} is the first benchmark for this task. We provide all the resources necessary for using and further extending the benchmark with improvements. Authors anticipate that this will inspire research in this direction by providing a way to measure and compare the performance of different approaches. 

\paragraph*{Reusability and Sustainability:}
There are two ongoing workshops related to KG generation from text, Text2KG\footnote{\url{https://aiisc.ai/text2kg2023/}} at ESWC and NLP4KGC\footnote{\url{https://sites.google.com/view/nlp4kg/}} at the Web Conference. Furthermore, there is a proposed special issue\footnote{\url{https://www.semantic-web-journal.net/blog/special-issue-knowledge-graph-generation-text}} on this theme at Semantic Web journal. This will be a useful resource for evaluating approaches presented in those venues. As the authors are also co-organizers of these events, they plan to maintain and provide improved future versions of the data in collaboration with those workshops. It's also important to note that the authors and organizations of the aforementioned workshops are not from a single organization but distributed across multiple organizations and making the proposed resource not dependent on a single organization. The code used to generate the resource is available making it possible for anyone to reproduce, improve or create derived work from it. 

\paragraph*{Resource Availability Statement:} Text2KGBench dataset is available from zenodo ~\footnote{\url{https://zenodo.org/record/7916716\#.ZFrX5ezML0r}}, and the code that is used to generate the benchmark, evaluation scripts, baselines, LLM outputs, evaluation results are available from Github\footnote{\url{https://github.com/cenguix/Text2KGBench}}. Raw datasets we used are TekGen corpus~\footnote{\url{https://paperswithcode.com/dataset/tekgen}} and WebNLG corpus\footnote{\url{https://gitlab.com/shimorina/webnlg-dataset/-/tree/master/release\_v3.0/en}}. The LLM models we used are LLaMA~\footnote{\url{https://github.com/juncongmoo/pyllama}} to derive Vicuna-13B\footnote{\url{https://github.com/lm-sys/FastChat}} and Alpaca-LoRA-13B\footnote{\url{https://github.com/tloen/alpaca-lora}}.For sentence similarity we used SBERT\footnote{\url{https://www.sbert.net/}} with T5-XXL model~\footnote{\url{https://huggingface.co/sentence-transformers/gtr-t5-xxl}}.

\bibliographystyle{unsrtnat}
\bibliography{iswc2023} 

\end{document}